\def\year{2022}\relax
\documentclass[letterpaper]{article} 
\usepackage{aaai22}  
\usepackage{times}  
\usepackage{helvet}  
\usepackage{courier}  
\usepackage[hyphens]{url}  
\usepackage{graphicx} 
\usepackage{natbib}  
\usepackage{caption} 
\DeclareCaptionStyle{ruled}{labelfont=normalfont,labelsep=colon,strut=off} 
\usepackage{algorithm}
\usepackage{algorithmic}

\usepackage{amsmath, amssymb}
\usepackage{bm}
\usepackage{physics}
\usepackage{color}
\usepackage{newfloat}
\usepackage{listings}
\floatstyle{ruled}
\newfloat{listing}{tb}{lst}{}
\floatname{listing}{Listing}
\title{Adversarial Bone Length Attack on Action Recognition}
\author{
     Nariki Tanaka,\textsuperscript{\rm 1}
     Hiroshi Kera,\textsuperscript{\rm 2}
     Kazuhiko Kawamoto,\textsuperscript{\rm 2}
 }
 \affiliations{
     \textsuperscript{\rm 1} Graduate School of Science and Engineering, Chiba University\\
     \textsuperscript{\rm 2} Graduate School of Engineering, Chiba University

     \{ntanaka, kera\}@chiba-u.jp, kawa@faculty.chiba-u.jp
 }

\usepackage{bibentry}

\begin{document}

\maketitle

\begin{abstract}
Skeleton-based action recognition models have recently been shown to be vulnerable to adversarial attacks. Compared to adversarial attacks on images, perturbations to skeletons are typically bounded to a lower dimension of approximately 100 per frame. This lower-dimensional setting makes it more difficult to generate imperceptible perturbations. Existing attacks resolve this by exploiting the temporal structure of the skeleton motion so that the perturbation dimension increases to thousands. 
In this paper, we show that adversarial attacks can be performed on skeleton-based action recognition models, even in a significantly low-dimensional setting without any temporal manipulation. Specifically, we restrict the perturbations to the lengths of the skeleton's bones, which allows an adversary to manipulate only approximately 30 effective dimensions. We conducted experiments on the NTU RGB+D and HDM05 datasets and demonstrate that the proposed attack successfully deceived models with sometimes greater than 90\% success rate by small perturbations. Furthermore, we discovered an interesting phenomenon: in our low-dimensional setting, the adversarial training with the bone length attack shares a similar property with data augmentation, and it not only improves the adversarial robustness but also improves the classification accuracy on the original data. This is an interesting counterexample of the trade-off between adversarial robustness and clean accuracy, which has been widely observed in studies on adversarial training in the high-dimensional regime.
\end{abstract}

\section{Introduction}
Deep neural network models are highly vulnerable to adversarial perturbations, which are small input perturbations intentionally applied by an attacker~\cite{Szegedy2014Intriguing}. This poses a security concern in the use of deep neural network models in practical scenarios. \citet{Szegedy2014Intriguing} was the first to discover the adversarial attack, that is, applying small perturbations to images that are imperceptible to a human but can fool deep neural network models. Since then, various adversarial attack methods have been proposed in computer vision~\cite{Goodfellow2015Explaining,Carlini2017Towards,Madry2018Towards}. 
Adversarial attacks are not limited to the image domain; they are also possible in domains such as video classification~\cite{Wei2019Sparse, Chen2021appending}, text classification~\cite{Fursov2021Differentiable}, and speaker recognition~\cite{Guangke2021Who}. 
\begin{figure}[t]
\includegraphics[width=\columnwidth]{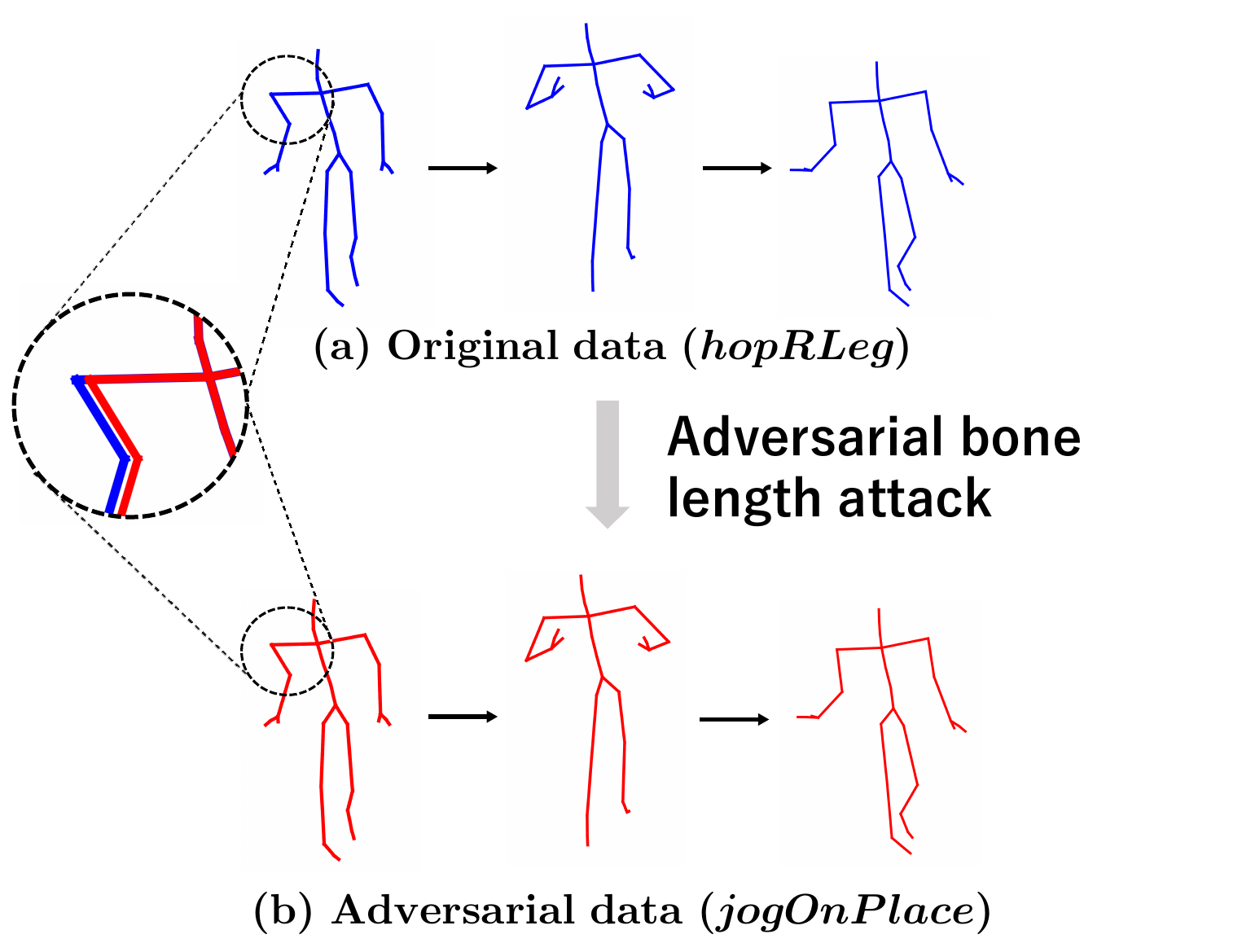}
\caption{The overview of the adversarial bone length attack. Only the bone length of a skeleton is perturbed (approximately 30 effective dimension). Nevertheless, the classification of the motion changes from (a) \textit{hopRLeg} to (b) \textit{jogOnPlace}. }
\label{fig:intro}
\end{figure}

One of the reasons underlying the current success of adversarial attacks in various domains and tasks is that adversarial attacks manipulate high-dimensional data~\cite{gilmer2018adversarial, simon2019first}. For example, in the image domain, even small datasets (e.g., CIFAR10~\citet{Krizhevsky2009LearningML}) have hundreds or thousands of pixels to perturb. In the video domain, more dimensions can be perturbed by exploiting the temporal structure. Recently, it was shown that adversarial attacks can succeed even on data in moderate-dimensional settings~\cite{Su_2019_one,Pony2021Over}. 
In particular, \cite{Liu2020Adversarial,Zheng2020Towards,Wang2021Understanding,Diao2021BASAR} considered adversarial attacks in skeleton-based action recognition. 
In this case, an attacker is allowed to perturb the shape and motion of the skeleton along frames, and a skeleton at each frame is represented by approximately 100 parameters (two or three-dimensional coordinates of approximately 30 joints). With constraints on bone connection, the effective number of parameters is even lower. In such a case, it is difficult for attackers to design adversarial perturbations that are sufficiently small to be imperceptible. In~\cite{Liu2020Adversarial,Zheng2020Towards,Wang2021Understanding,Diao2021BASAR}, because the motion of the skeleton along frames also was perturbed, they achieved a high success rate of attack with imperceptible adversarial perturbations. 

In this study, we consider an extremely low-dimensional adversarial attack on skeleton-based action recognition, where only the lengths of the skeleton's bones can be perturbed, as shown in Fig.~\ref{fig:intro}.
The proposed attack has access to only approximately 30 dimensions, which is significantly lower than existing attacks. 
Despite the restrictiveness of our new attack setting, we experimentally observed that such an extremely low-dimensional attack can succeed. 
Our bone length attack only requires people to change the apparent bone lengths, e.g., by covering some parts of the body with clothes or attaching a fake extension to arms. 
In contrast, 
existing skeleton-based methods require people to move all the joint positions adversarially, which is almost infeasible because perfect body coordination is required all along the motion.
Our experiments were conducted on two datasets, the NTU RGB+D~\cite{Shahroudy2016NTU} and HDM05~\cite{M2007Documentation} datasets, and two models, the spatial-temporal graph convolutional network~(ST-GCN;~\citet{Yan2018Spatial}) and semantics-guided neural network~(SGN;~\citet{Zhang2020Semantics}). In such settings, we saw that with a small perturbation, our attack successfully fooled the models with a 20\% success rate, reaching 90\% in some cases.
We also investigated which parts of a skeleton are more susceptible to our bone length adversarial attack and discovered that perturbing the bones that are long and close to the root joint (base of spine) is effective. 
Another interesting observation is that adversarial training improves not only the robustness against our attack but also the accuracy of the original data. In the literature~\cite{zhang2019theoretically}, it has been established that adversarial training improves the robustness against adversarial examples at the cost of accuracy in original data. Our observations provide a counterexample of this. Based on our experimental results, we consider this to be because in our low-dimensional setting, data augmentation and adversarial examples have similar properties.

Due to the broad and important downstream applications of skeleton-based action recognition, it is crucial to consider the potential vulnerability of skeleton-based action recognition against adversarial attacks. In particular, our adversarial bone length attack seems more realistic than others because it only requires  deception of a skeleton extractor to wrongly measure the length of bones, whereas other attacks require perturbation of all positions of joints as well as their time evolution along frames. 

Our contributions can be summarized as follows:
\begin{itemize}
    \item We propose the first adversarial attack that only perturbs the lengths of bones to fool skeleton-based action recognition models. Unlike existing adversarial attacks, our attack works in an extremely low-dimensional setting (approximately 30 dimensions and no temporal perturbation). 
    \item Through extensive experiments, we demonstrate the effectiveness of our bone length attack. We also discovered that bones that are long and close to the base of the spine are significantly more susceptible to the attack than other bones.
    \item We discovered that adversarial training using the proposed attack improves not only the adversarial robustness against this attack but also the accuracy against the original data. We also observed that data augmentation improves both the clean accuracy and adversarial robustness.
    This implies that a low-dimensional adversarial attack may have distinct characteristics from other adversarial attacks in high-dimensional settings. 
\end{itemize}

\section{Related Work}
\paragraph{Skeleton-based action recognition.} Skeleton-based action recognition has attracted significant attention in recent years. There are many advantages to using skeleton data for action recognition. Skeleton data are considered to be robust to lighting, subject clothing, and background~\cite{sun2021human,Wang2021Understanding}. They are also superior to RGB data in terms of computational cost. Due to these advantages and the progress of sensors~\cite{Zhengyou2012Kinect} and pose estimation models~\cite{Wandt2021CanonPose, Xu2021Graph}, various models have been proposed~\cite{Yan2018Spatial,Zhang2020Semantics,Ke2021Extremely,Jun2021Symmetrical}. 
\paragraph{Adversarial attacks on skeletons.} Adversarial attacks in skeleton-based action recognition have been proposed very recently~\cite{Liu2020Adversarial,Zheng2020Towards,Wang2021Understanding,Diao2021BASAR}. 
\citet{Liu2020Adversarial} was the first to propose an attack on this task. Their proposed attack generates natural motions satisfying multiple physical constraints by fixing the bone length, limiting the perturbation magnitude and joint acceleration, and using a generative adversarial network~\cite{Goodfellow2014Generative}. 
They also showed that adversarial examples remain adversarial even after being converted from skeleton to RGB data and then from RGB to skeleton data.
\citet{Zheng2020Towards} proposed an attack by restricting the change in angle between bones. They also proposed a defense method. \citet{Wang2021Understanding} conducted user studies to show that their attack is highly imperceptible to humans. They also argued that joints with high velocity and acceleration are vulnerable features. These attacks assume that the attacker has complete knowledge of the model being attacked, such as its parameters and structure, and are called white box attacks. In contrast, \citet{Diao2021BASAR} proposed a black box attack, which assumes complete ignorance of the information about an attacked model. Most of the aforementioned adversarial attacks increase the effective dimensions of the attack to a few thousand by using degrees of freedom in the temporal direction. In contrast, we propose an adversarial attack that has no freedom in the temporal direction. This restriction makes this study largely different from  previous studies.

\paragraph{Low-dimensional adversarial attacks.}
Our proposed attack perturbs only approximately 30 dimensions. There exist several related attacks. In the image classification domain, \citet{Su_2019_one} proposed a one-pixel attack, which perturbs only a single pixel. However, attackers can select the target pixel (three channels) from a large number of pixels; thus, the effective dimension is still higher than ours. In addition, a one-pixel attack is relatively easily noticeable. In the video classification domain, the attack proposed by~\cite{Pony2021Over} does not use any spatial information, and thus the available dimensions are low. However, the available dimensions in our attack are less than those in their attack.
Moreover, neither \citet{Su_2019_one} nor \citet{Pony2021Over} explored the effect of adversarial training, while we observed an interesting result through adversarial training.

\section{Method}
In this section, we explain how to attack skeleton-based action recognition models by perturbing the bone length only.

\begin{figure}[t]
\begin{center}
\includegraphics[width=0.80\columnwidth]{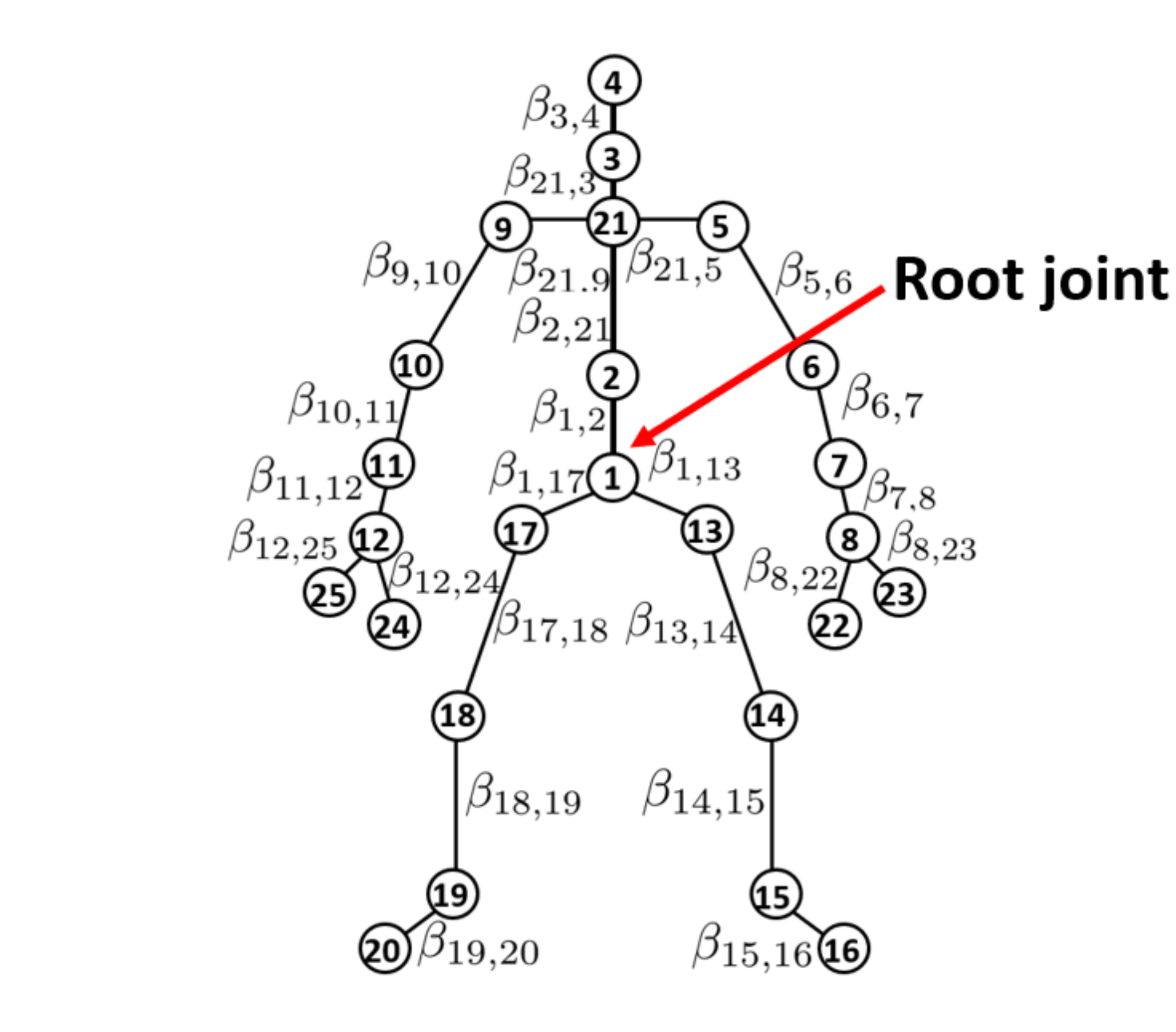}
\end{center}
\caption{The skeleton of the NTU RGB+D dataset. This skeleton consists of 24 bones, and each joint is indexed.
We define the base of the spine (joint 1) as the root joint. Our adversarial bone length attack perturbs the scale parameter $\qty{\beta_{i,j}}$.
}
\label{fig:skeleton}
\end{figure}

\subsection{Notations}
We consider $L$-class skeleton-based action recognition. 
An action (sequence of skeletons) is represented by $X=\qty{ \bm{q}_i(t) \mid i=1,\ldots,M, t=0,\ldots,T-1}$, where $T$ and $M$ denote the numbers of frames and joints, respectively, and $\bm{q}_i(t)\in \mathbb{R}^3$ denotes the 3D coordinates of the $i$-th joint of the skeleton at frame $t$. A classifier $f$ receives an action $X$ and outputs an $L$-dimensional confidence vector, whose $k$-th entry, denoted by $f(X)_k$, represents the confidence for the $k$-th class. A class label is encoded in a one-hot representation as $\bm{y} = (y_1, \ldots, y_L)^{\top}$. 

\subsection{Bone length parameters}\label{subsec:bone length-parameters}

Our attack perturbs the lengths of bones. 
The length of the $(i,j)$-th bone, which connects the $i$-th and $j$-th joints, is associated with a scale parameter $\beta_{i,j}$, as shown in Fig.~\ref{fig:skeleton}. Note that simply perturbing the length scale of each bone does not directly yield a valid skeleton. Therefore, the length scales $\{\beta_{i,j}\}$ of bones are perturbed from the root joint sequentially as follows.  
First, we define the base of the spine as the root joint $\bm{q}_{\mathrm{root}}(t)$. 
Then, starting from the root joint, the position of each joint is set sequentially. Specifically, when a joint $\bm{q}_i(t)$ is perturbed to $\widetilde{\bm{q}}_i(t)$, each of its child joints (say, $\bm{q}_j(t)$) is perturbed as follows: 
\begin{align} \label{eq:param}
\widetilde{\bm{q}}_j(t) &= \beta_{i,j}(\bm{q}_j(t)-\bm{q}_i(t))+\widetilde{\bm{q}}_i(t).
\end{align}
Note that the root is fixed, i.e., $\widetilde{\bm{q}}_{\mathrm{root}}(t)=\bm{q}_{\mathrm{root}}(t)$. 
As can be seen in Eq.~(\ref{eq:param}), the value of $\beta_{i,j}$ is the ratio between the length of the $(i,j)$-th bone before and after the adversarial perturbation.
After the positional rearrangement by Eq.~(\ref{eq:param}), the final adversarial example of $X$ is given by $\widetilde{X}=\qty{\widetilde{\bm{q}}_i(t) \mid i=1,\ldots,M, t=0,\ldots,T-1}$. 

\begin{algorithm}[t]
\caption{Pseudocode of adversarial bone length attack}
\label{alg:attack}
\textbf{Input}: Original sequence of skeleton $X$, ground truth label $y$, trained classifier $f$\\
\textbf{Parameter}: Step size $\alpha$, maximum iteration number $N$, bound $\epsilon$\\
\textbf{Output}: Adversarial example $\widetilde{X}$
\begin{algorithmic}[1] 
\STATE $\bm{\beta}^{(0)}\leftarrow \bm{1}$
\FOR{$n\leftarrow 0$ to $N-1$}
\STATE $\widetilde{X}\leftarrow\mathrm{SetParam}\qty(\bm{\beta}^{(n)},X)$
\\$\triangleright~\mathrm{SetParam}(\cdot,\cdot)$ outputs an adversarial example $\widetilde{X}$ consisting of $\widetilde{\bm{q}}_j(t)$ defined using $\bm{\beta}^{(n)}$ in Eq.~(\ref{eq:param})
\STATE $\bm{z} \leftarrow f(\widetilde{X})$\\
$\triangleright~\bm{z}$ is the confidence vector, which is output by $f$ that received $\widetilde{X}$.
\STATE $y_{\mathrm{pred}} \leftarrow \mathrm{Predict}\qty(\bm{z})$ \\$\triangleright~\mathrm{Predict}(\cdot)$ outputs the class label predicted by the classifier $f$
\IF {$y_{\mathrm{pred}}\neq y$}
\STATE \textbf{break}
\ENDIF
\STATE $\bm{\beta}^{(n+1)}\leftarrow\mathrm{Clip}_{\bm{\beta}^{(0)}, \epsilon}\qty(\bm{\beta}^{(n)}+\alpha\cdot \mathrm{sign}(\nabla_{\bm{\beta}^{(n)}}\mathcal{L}\qty(\bm{\beta}^{(n)})))$
\\$\triangleright~\mathcal{L}(\cdot)$ is cross entropy loss
\IF {$n=N-1$}
\STATE $\widetilde{X}\leftarrow\mathrm{SetParam}\qty(\bm{\beta}^{(N)},X)$
\ENDIF
\ENDFOR
\STATE \textbf{return} $\widetilde{X}$
\end{algorithmic}
\end{algorithm}

\begin{figure}[t]
\begin{center}
\includegraphics[width=\columnwidth]{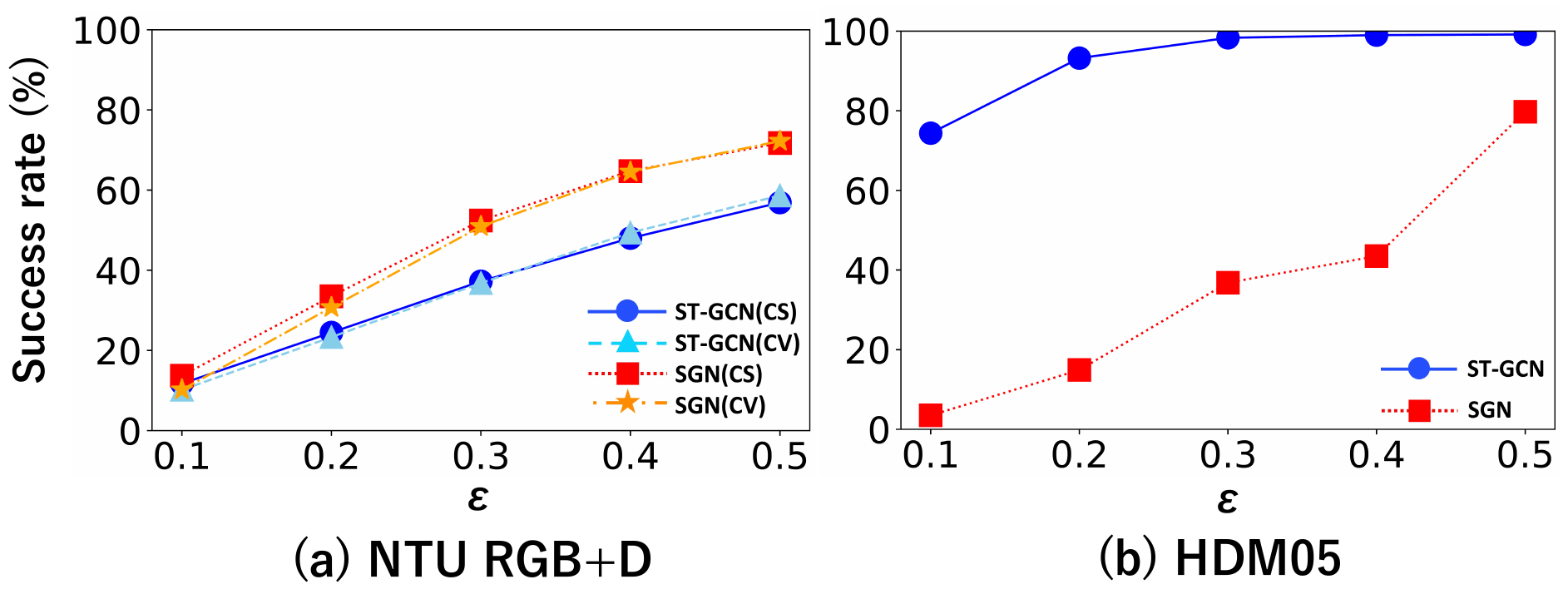}
\end{center}
\caption{Success rates of the adversarial bone length attack on the ST-GCN and SGN models according to attack strength $\epsilon$: (a) NTU RGB+D; two configurations of the training sets, cross-subject (CS) and cross-view (CV), were considered. Similar results were observed across models and datasets. The SGN model was slightly more vulnerable than the ST-GCN model. (b) HDM05; The ST-GCN model was extremely vulnerable, even for small perturbations.}
\label{fig:result1}
\end{figure}

\subsection{Adversarial bone length attack}
Our attack is based on the following optimization with respect to the bone length parameters $\qty{\beta_{i,j}}$. Let $\bm{\beta}$ be a vector that lists the elements of $\qty{\beta_{i,j}}$. Then, the adversarial scales of bones are obtained through the following cross-entropy loss maximization:
\begin{align}
&\max_{\bm{\beta}}\ \mathcal{L}(\bm{\beta}) = -\sum_{k=1}^L y_k \log f(\widetilde{X})_k,\
 \text{s.t. } \|{\bm{\beta}-\bm{\beta}^{(0)}}\|_{\infty} \leq \epsilon, 
\end{align}
where $f$ is the target classifier that we attempt to attack. The initial value of $\bm{\beta}$ is set to an all-one vector, i.e., $\bm{\beta}^{(0)} = \bm{1}$. The magnitude of the perturbation is bounded by $\epsilon$. We optimize $\bm{\beta}$ using projected gradient descent (PGD) (Eq.~(\ref{eq:pgd}))~\cite{Madry2018Towards}. The parameter $\bm{\beta}$ is iteratively updated by
\begin{align} \label{eq:pgd}
\bm{\beta}^{(n+1)}&=\mathrm{Clip}_{(\bm{\beta}^{(0)}, \epsilon)}\qty[\bm{\beta}^{(n)}+\alpha\cdot \mathrm{sign}(\nabla_{\bm{\beta}^{(n)}}\mathcal{L}\qty(\bm{\beta}^{(n)}))],
\end{align}
where $\alpha > 0$ is the step size, and $\bm{\beta}^{(n)}$ is the $\bm{\beta}$ obtained after $n$ iterations. $\mathrm{Clip}_{(\bm{\beta}^{(0)}, \epsilon)}\qty[\cdot]$ is an operator that clips each entry of a given vector to $[1-\epsilon, 1+\epsilon]$, and $\mathrm{sign}(\cdot)$ denotes the sign function, which maps each element of the argument vector to $\pm 1$ according to its sign. 

\begin{table}[t]
\begin{tabular}{|c|c|c|c|c|c|}
\hline
 &$\epsilon=0.1$&$\epsilon=0.2$&$\epsilon=0.3$&$\epsilon=0.4$&$\epsilon=0.5$\\ \hline
ES &51.3\%&50.6\%&50.4\%&50.4\%&50.4\% \\\hline
FR &74.6\%&81.4\%&83.3\%&83.1\%&82.9\% \\\hline
\end{tabular}
\caption{Average of the confidence scores given by the ST-GCN model for misclassified adversarial examples on the HDM05 dataset. The average confidence in early-stopping case~(ES) was low, while it was high in full-run case~(FR).
}
\label{tab:conf}
\end{table}

The pseudo code of our attack is given in Algorithm~\ref{alg:attack}.
First, the bone scale parameter $\bm{\beta}$ and adversarial example $\widetilde{X}$ are initialized as the all-one vector and $X$, respectively. Then, the following procedure is repeated.
\begin{enumerate}
    \item $\widetilde{X}$ is input to the target classifier. If the target classifier misclassifies, the attack terminates; otherwise, $\bm{\beta}$ is updated according to Eq.~(\ref{eq:pgd}).
    \item The action $\widetilde{X}$ is updated according to Eq.~(\ref{eq:param}).
\end{enumerate}
 This termination condition of the attack is according to the official code provided by \cite{Wang2021Understanding}.
 In the experiments, we also restricted our attack to a subset of bones to examine which part of skeleton is the most sensitive to the perturbation. In this case, only the subset of $\{\beta_{i,j}\}$ was considered in Eq.~(\ref{eq:pgd}). We also attacked the classifier using Adam~\cite{Kingma2015Adam} optimizer instead of PGD, as was used in other studies for attacking skeleton-based action recognition models~\cite{Liu2020Adversarial,Wang2021Understanding}. In this case, the update rule in Eq.~$(\ref{eq:pgd})$ was replaced with that of Adam.

\begin{figure*}[t]
\begin{center}
\includegraphics[width=1.0\textwidth]{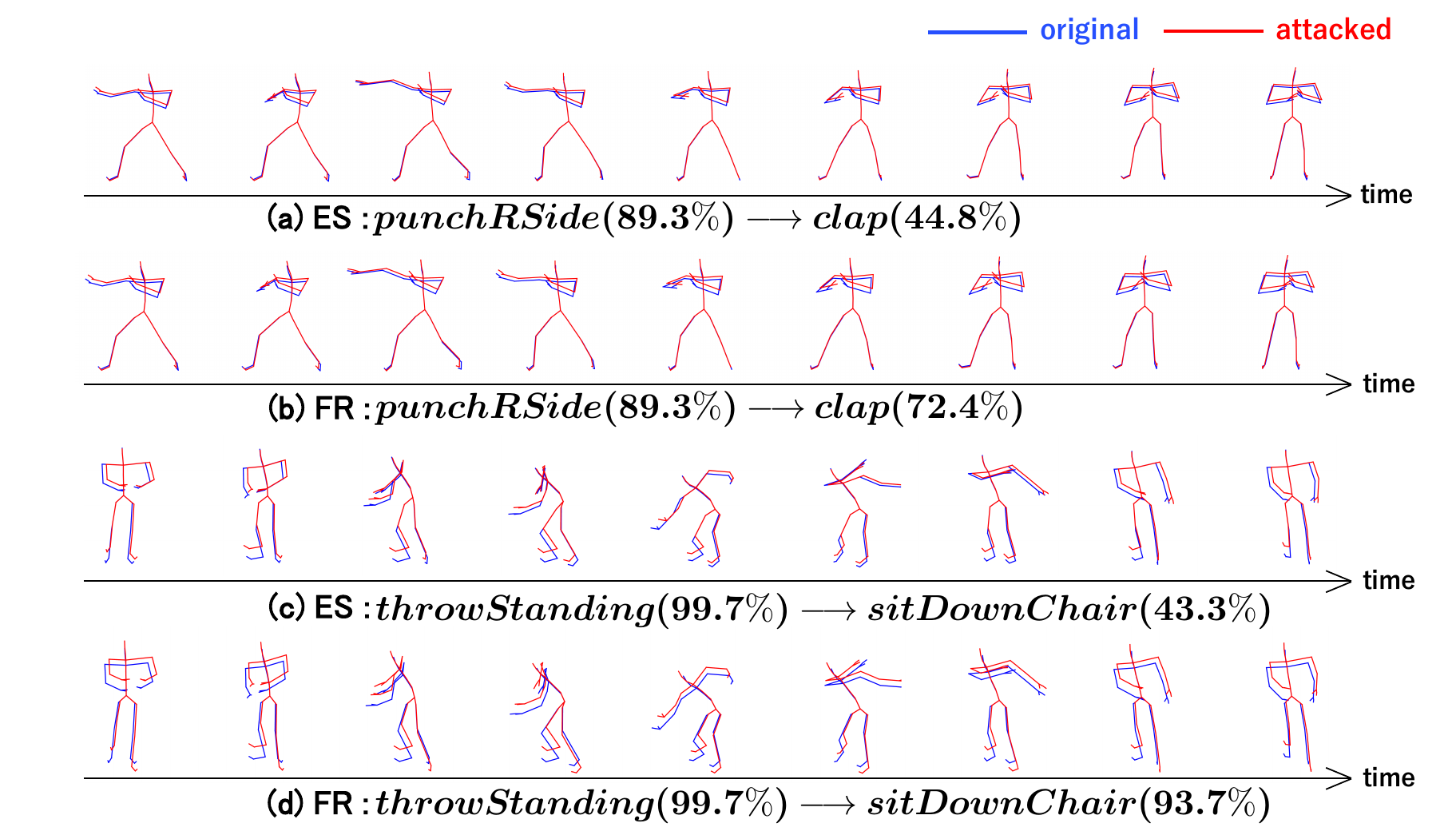}
\end{center}
\caption{Motion of skeletons before and after adversarial perturbations (blue and red, respectively) when we attacked the ST-GCN model using the HDM05 dataset. In the early-stopping~(ES) case, we terminated the iteration in attack once the adversarial example succeeded in fooling the model. In the full-run~(FR) case, the adversarial example was updated for a fixed number of 50 iterations. (a, b) the original motion \textit{punchRSide} turned into adversarial motion \textit{clap}. With ES, there is almost no visible change in the skeleton but the prediction confidence remains moderate, whereas with FR, the change becomes slightly more visible and the prediction confidence is high.
(c, d) the original motion \textit{throwStanding} turned into adversarial motion \textit{sitDownChair}. The results are similar to those observed in (a,b).}
\label{fig:examples}
\end{figure*}

\section{Experiments and Results}
We first introduce our experimental settings. Then, we present the results of our proposed attack. Finally, we attempt to defend against our attack.
\subsection{Experimental settings}
\label{subsec:settings}
\paragraph{Datasets}
We used the NTU RGB+D~\cite{Shahroudy2016NTU} and HDM05~\cite{M2007Documentation} datasets, which are 3D skeleton action datasets.
The NTU RGB+D dataset consists of 56,880 motion data of a skeleton with 24 bones.  This dataset has 60 classes and was created from 40 subjects. There are two ways to split the training and test data: cross-subject (CS) and cross-view (CV). In CS, the dataset is divided so that the numbers of subjects included in the training and test data are both 20. The numbers of samples are 40,320 and 16,560, respectively. CV divides the training and test data according to the camera's viewpoint. The numbers of samples in these training and test datasets are 37,920 and 18,960, respectively.
The HDM05 dataset consists of 2,337 data of a skeleton with 30 bones. The number of classes is 130. 
At the prepossessing stage, we adopted a smoothing filter, translation, and so on. This reduced the number of classes to 65. We randomly divided samples of each class into a training set~(80\%), validation set~(10\%), and testing set~(10\%). Please refer to the supplementary material for this preprocess. 

\paragraph{Target models}
We used two skeleton-based action recognition models as our target models to attack: the ST-GCN~\cite{Yan2018Spatial} and SGN~\cite{Zhang2020Semantics} models. When the NTU RGB+D dataset was used, we used pretrained weights provided by the authors of each model\footnote{https://github.com/yysijie/st-gcn}\footnote{https://github.com/microsoft/SGN}. When the HDM05 dataset was used, we trained the two models with the official code provided by the authors. To guarantee the convergence, we ran at least 300 epochs for the ST-GCN model and 200 epochs for the SGN model and adopted early stopping for both models. The data augmentation for each model was the same as when the model was trained on the NTU RGB+D dataset. After training, the test accuracies on the ST-GCN and SGN models were 86.1\% and 96.1\%, respectively. 

\paragraph{Evaluation metrics and others}
Following~\cite{Wang2021Understanding}, we attacked the test data that were correctly classified by the target model. Then, the attacks were evaluated based on their success rate. In this paper, we used PGD~(Eq.~\ref{eq:pgd}). We leave the results using Adam in the supplementary material. The maximum number of iterations of the PGD was set to 50. The step size was set to $\alpha=0.01$, as in~\cite{Liu2020Adversarial}. 
All experiments were conducted using an Intel Core i7-6850K CPU and TITAN RTX GPU. 

\subsection{Attack results}

To investigate the effectiveness and imperceptibility of our attack,  we attacked two models. In  Fig.~\ref{fig:result1} (a), we show the results of the attack using the NTU RGB+D dataset. As a result of the attack on the ST-GCN model, the success rates for the PDG attack with $\epsilon=0.1$ and $\epsilon=0.2$ exceeded 10\% and 20\%, respectively. Note that this success rate is not as high as those achieved in other studies; this is because the dimensionality of the input data used for the attack in these studies was much higher (e.g., thousands of dimensions), whereas in our setting, the dimensionality is only approximately 30. Nevertheless, our attack achieved moderate success rates that were nonnegligible for practical use. When we set $\epsilon=0.5$, we can see that the attack became more successful~(greater than $50\%$ success). For the HDM05 dataset, the success rate was even higher~(Fig.~\ref{fig:result1} (b)). We observed that the success rate of the attack on the ST-GCN model exceeded  $70\%$ when $\epsilon=0.1$. With $\epsilon=0.2$, the success rate was over $90\%$.

The results indicate that the proposed attack is very effective for some models and datasets. In Fig.~\ref{fig:examples} (a),  we provide an adversarial example generated with $\epsilon=0.1$ when we attacked the ST-GCN model using the HDM05 dataset. One can see that the predicted classes are different although the skeletons appear very similar before and after the attack (blue and red, respectively). The same result was observed for $\epsilon=0.2$~(Fig.~\ref{fig:examples} (c)). To summarize, the ST-GCN model trained by the HDM05 dataset was very vulnerable to our bone length attack on the skeleton.
However, the confidences on the adversarial motions in Figs.~\ref{fig:examples} (a) and (c) are relatively low, because we employ early-stopping (ES); the proposed attack terminates once the adversarial example fools the target model \cite{Wang2021Understanding}. To make the confidences higher, we can increase the number of iterations in the PGD attack.
Hence we attacked the model with 50 iterations, which is the maximum number of iterations in the PGD attack.
We call the termination condition the {\it full-run}  (FR) case.
Table~\ref{tab:conf} shows that the average confidence scores in the ES and FR cases for misclassified adversarial examples and that the confidence scores in the FR case are higher than those in the ES case.  In Figs.~\ref{fig:examples} (b) and (d), we demonstrate the FR attacks using the same original data in Figs.~\ref{fig:examples} (a) and (c), respectively.
From Figs.~\ref{fig:examples}, 
one can see the FR attacks cause bigger changes than the ES attacks.
These results come from that there is a trade-off between high confidence and imperceptibility.

\begin{figure}[t]
\begin{center}
\includegraphics[width=1\columnwidth]{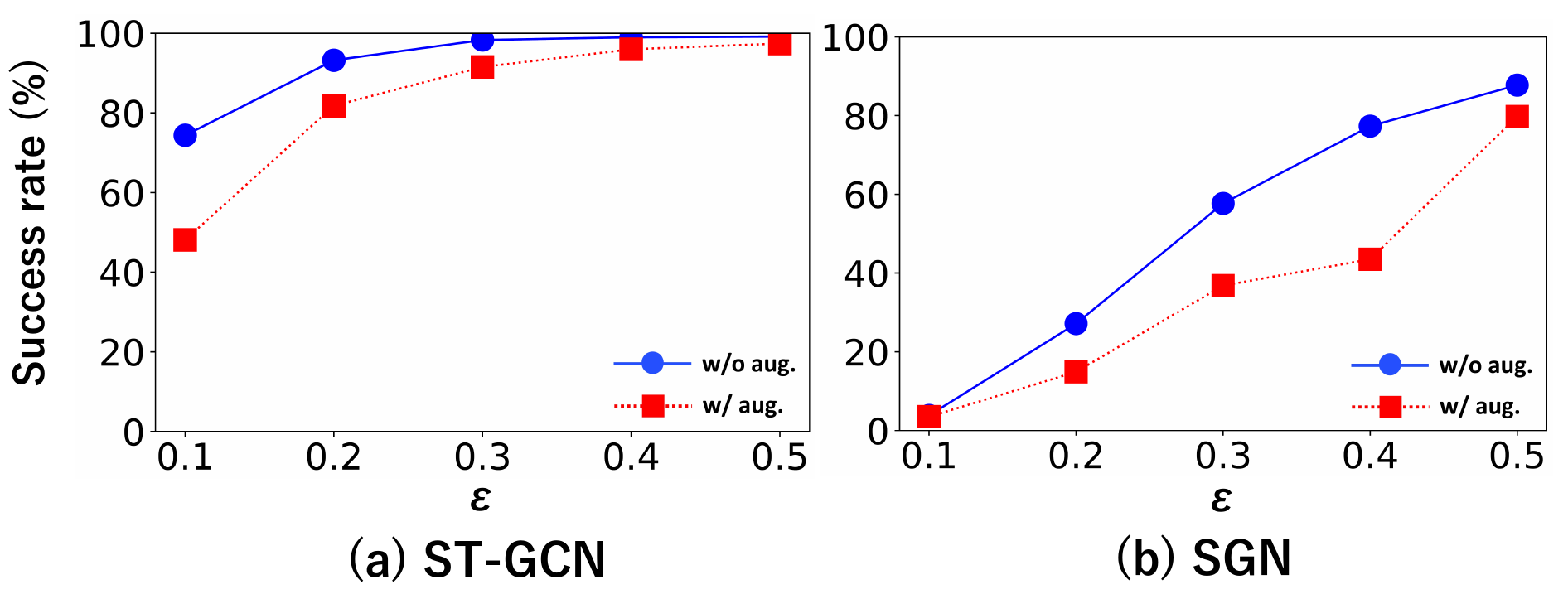}
\end{center}
\caption{Success rates of the adversarial bone length attack for the (a) ST-GCN and (b) SGN models on the HDM05 dataset with and without data augmentation. 
}
\label{fig:result2}
\end{figure}

\begin{table}[t]
\begin{center}
\begin{tabular}{|c|c|c|c|}
\hline
Model&w/o aug. &w/ aug.\\ \hline
ST-GCN&86.1\%&89.8\% \\\hline
SGN&95.3\%&96.1\% \\\hline
\end{tabular}
\caption{Classification accuracy of models trained with the HDM05 dataset with and without data augmentation (\textit{w/ aug} and \textit{w/o aug}, respectively). One can see that data augmentation improved the accuracy for both models.}
\label{tab:acc}
\end{center}
\end{table}

\begin{figure}[t]
\begin{center}
\includegraphics[width=0.6\columnwidth]{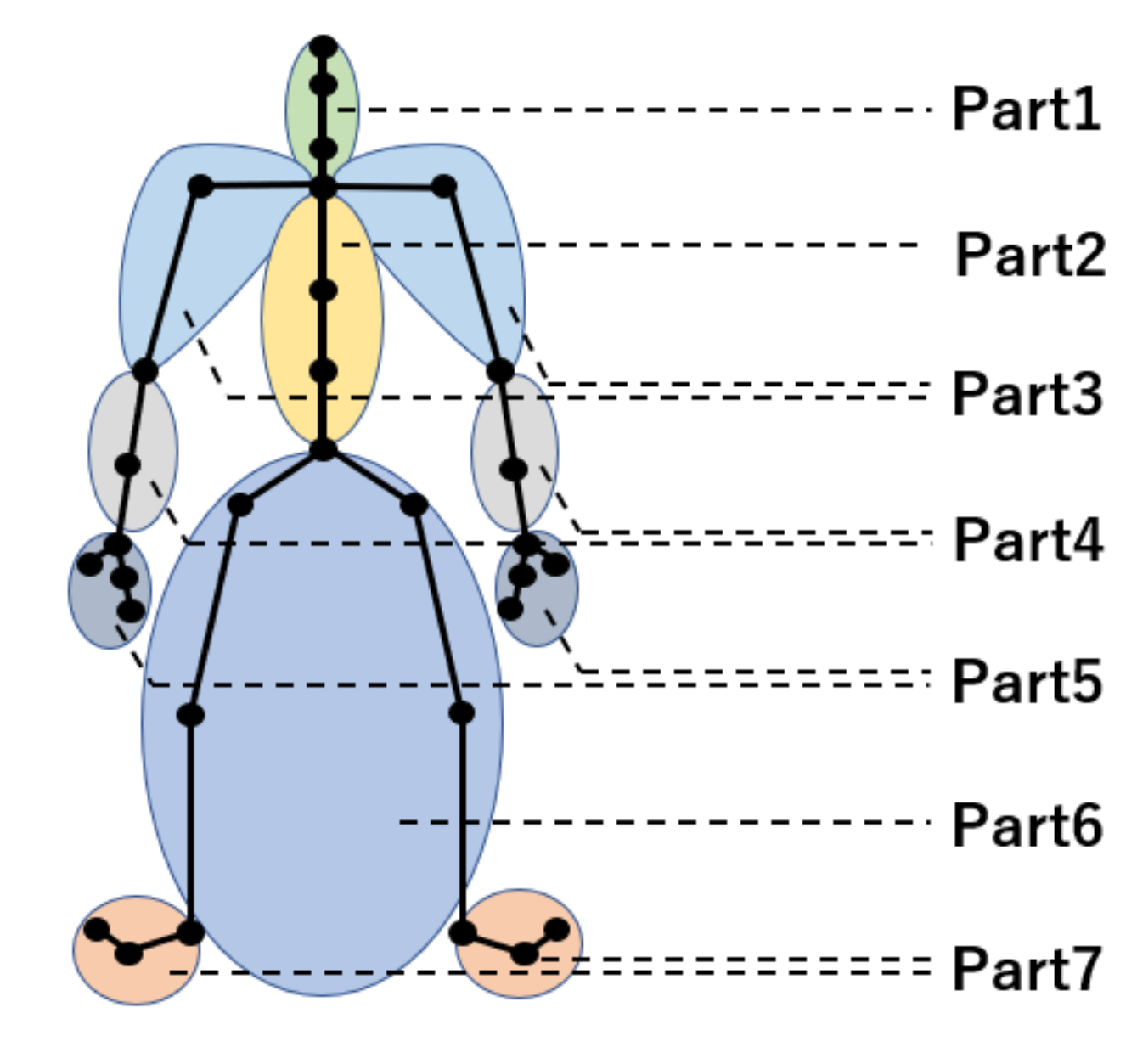}
\end{center}
\caption{Grouping of bones.  The skeleton bones were divided into seven symmetrical parts.}
\label{fig:partition}
\end{figure}
\begin{table}[t]
\begin{center}
\begin{tabular}{|c|r|r|r|}
\hline
Part&$\epsilon=0.1$&$\epsilon=0.2$&$\epsilon=0.3$\\ \hline
Part 1&0.51\%&1.70\%&2.20\% \\\hline
Part 2&12.6\%&28.9\%&40.7\% \\\hline
Part 3&9.34\%&25.1\%&41.4\% \\\hline
Part 4&2.38\%&5.94\%&11.9\% \\\hline
Part 5&0.00\%&0.51\%&0.85\% \\\hline
Part 6&34.1\%&67.2\%&75.8\% \\\hline
Part 7&0.00\%&0.68\%&1.35\% \\\hline
\end{tabular}
\caption{Success rates of adversarial bone length attack on the ST-GCN model with the HDM05 dataset restricted to different parts. 
The sensitivity to the attack was strongly dependent on the parts.
In particular, Part~6 was the most sensitive to perturbations. }
\label{tab:part}
\end{center}
\begin{center}
\begin{tabular}{|c|c|c|c|}
\hline
Part&$\epsilon=0.1$&$\epsilon=0.2$&$\epsilon=0.3$\\ \hline
Part3+Part4+Part5&14.4\%&39.0\%&59.6\% \\\hline
\end{tabular}
\caption{Success rates of adversarial bone length attack on the ST-GCN model with the HDM05 dataset by restricted to the union of Parts~3,~4,~and~5, which contains more total number of bones as Part~6. The sensitivity to the attack in this region was lower than that in Part~6~(Table~\ref{tab:part}).
}
\label{tab:attack_3p}
\end{center}
\end{table}

As we can see the results for the HDM05 dataset shown in Fig.~\ref{fig:result1}(b), attacks on the two models resulted in significantly different success rates; the ST-GCN model was highly vulnerable to the attack, while the SGN model was more adversary-robust. 
We believe the reason underlying the adversarial robustness of the ST-GCN model lies in the data augmentation process. When we trained the ST-GCN and SGN models with the HDM05 dataset, we followed the training protocol with the NTU RGB+D dataset from the corresponding papers of the two models. Therefore, we did not use data augmentation when the ST-GCN model was trained with the NTU RGB+D dataset, but we did for the SGN model (Sec.~\ref{subsec:settings}).
To observe the effect of data augmentation on adversarial robustness, we trained the models on the HDM05 dataset with and without data augmentation and attacked them. 
The data augmentation during training of the SGN model was performed by randomly rotating the skeleton. We adopted this augmentation for the ST-GCN model as well because we did not want the results to depend on the quality of the data augmentation. 
As shown in Table~\ref{tab:acc}, the classification accuracy of both models for original data was improved by data augmentation. Interestingly, as shown in Fig.~\ref{fig:result2}, data augmentation also improved the adversarial robustness (lower success rates). In particular, the improvement in adversarial robustness for the ST-GCN model was significant~(approximately 20\% improvement at $\epsilon=0.1$). Therefore, the adversarial robustness exhibited by the ST-GCN model in Fig.~\ref{fig:result1}(b) can be attributed to the presence of data augmentation.

In image classification, some data augmentation methods were proposed to improve adversarial robustness~\cite{zhang2018mixup, yun2019cutmix}. However, in white box setting, these methods are effective only for non-iterative attack e.g., FGSM~\cite{Goodfellow2015Explaining}.
To the best of our knowledge, the literature has not previously reported data augmentation significantly improving adversarial robustness against iterative attack. In \cite{rice2020overfitting} and \cite{gowal2020uncovering}, adversarial training was combined with various data augmentations, but no particular improvement in adversarial robustness was observed.
We hypothesize that in the low-dimensional setting of this study, adversarial examples and augmented data are much more similar than in a high-dimensional setting, where adversarial examples are usually considered. We show promising empirical results regarding this hypothesis in Sec.~\ref{subsec:adversarial-training-experiment}.

Next, we investigated which subsets of bones were more vulnerable to perturbations. For this, we used the ST-GCN model and the HDM05 dataset. The skeleton was divided into seven parts (Fig.~\ref{fig:partition}), and adversarial perturbation was only applied to one of them. The results are summarized in Table~\ref{tab:part}. One can see that attacks on Parts~1,~4,~5,~and~7~(head, wrists, hands, and feet, respectively) almost completely failed, whereas attacks on Parts~2,~3,~and~6~(body, shoulders, and legs, respectively) were successful. In particular, the success rate for Part 6 was significant compared to those for the other parts. Thus, the legs are more susceptible to attacks. 
Part 6 has more bones than the other parts. To determine if the number of bones causes the vulnerability, we considered a joint attack on Parts~3,~4,~and~5, which have more total bones than Part 6 alone. The results are presented in Table \ref{tab:attack_3p}. It can be seen that the attack adding perturbations to Parts~3,~4, and~5 resulted in a much lower success rate. As such, the vulnerability of Part 6 is not due to its number of bones. 

Based on the above results, we found two tendencies. First, the closer the perturbed bones are to the root joint, the more successful the attack is likely to be. We believe that because a change in bone length can move the positions of the bones of their descendants, a change in bones closer to the root joint can have a greater impact on the entire skeleton. Second, the longer the perturbed bones are, the more likely the attack will be successful. The $\epsilon$ in this attack is proportional to the original bone. Therefore, for the same $\epsilon$, the longer the perturbed bone is, the greater the amount of bone change is allowed. These results differ from those of~\cite{Wang2021Understanding}. They stated that joints with greater velocity and acceleration are more useful for attacks. In contrast, our results show that attacks that perturb the torso are more successful than those that perturb the hands and arms. Therefore, we can see that our attack has different characteristics to theirs.

\subsection{Adversarial Training}\label{subsec:adversarial-training-experiment}

\begin{figure}[t]
\begin{center}
\includegraphics[width=1\columnwidth]{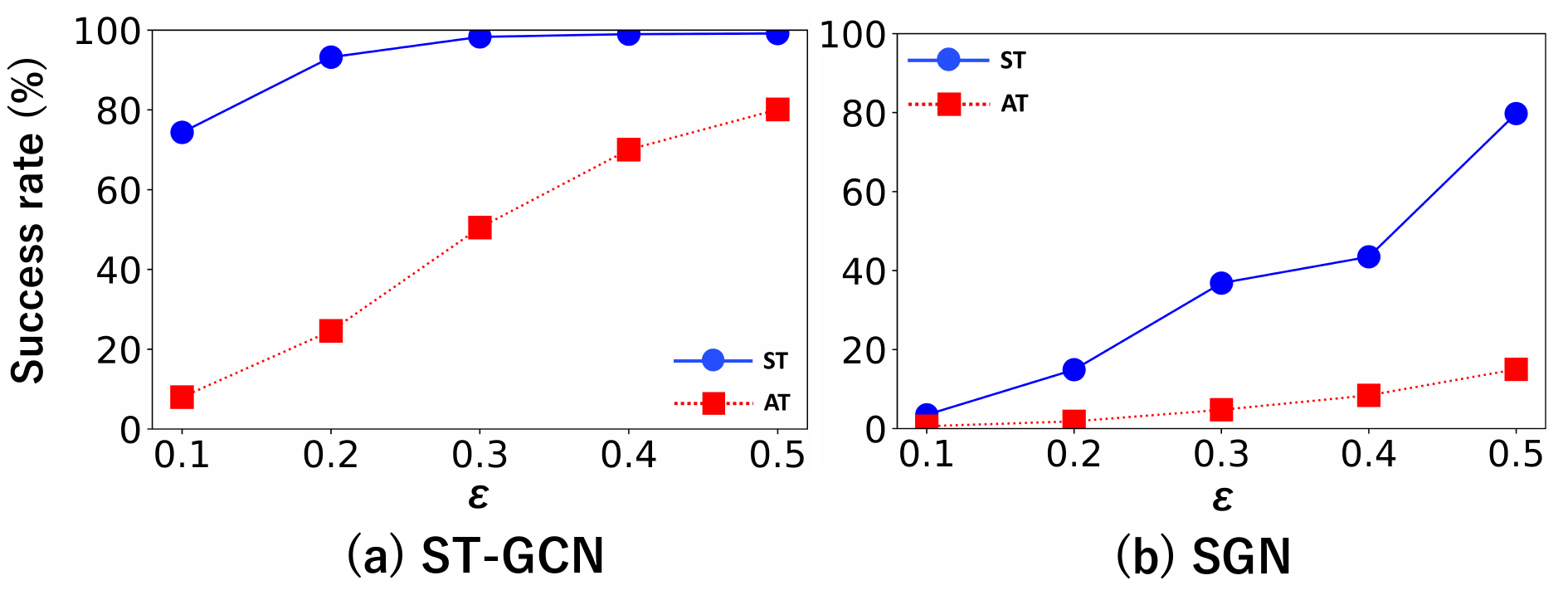}
\end{center}
\caption{Success rates of the adversarial bone length attack on models with standard training~(ST) and adversarial training~(AT).}
\label{fig:at_result}
\end{figure}

Next, we attempt to defend against our attack by performing adversarial training~\cite{Madry2018Towards} on the HDM05 dataset. We used our attack with $\epsilon=0.1$ in adversarial training.
The results are shown in Fig.~\ref{fig:at_result}. 
One can see that the ST-GCN and SGN models with adversarial training were more robust than that trained using the original data. By adversarial training, the ST-GCN model acquired the same level of robustness against our attack as the SGN model, and the SGN model provided only a low success rate even for large perturbation $\epsilon$. These results suggest that the proposed attack can be prevented to some extent by adversarial training.
Notably, we can also see that adversarial training increased the accuracy on the original data, as shown in ~Table~\ref{tab:at_acc}. The clean accuracy of the ST-GCN model was 86.1\% with standard training and 89.8\% with adversarial training. This phenomenon is a counterexample of the following widely seen observation in the literature: adversarial training gains adversarial robustness at the cost of reduced clean accuracy. We speculate that there may be a relationship between our attack and data augmentation as a factor in this phenomenon. As shown in Fig.~\ref{fig:result2}, we achieved some adversarial robustness against our attack by data augmentation. Data augmentation was used to increase the accuracy of the model, similar to our results as shown in Table.~\ref{tab:acc}, and adversarial training with our attack also increased the accuracy. In other words, adversarial training with our attack has similar properties to data augmentation. The details of this will be the subject of future work.
\begin{table}[t]
\begin{center}
\begin{tabular}{|c|c|c|c|}
\hline
Model&ST &AT\\ \hline
ST-GCN&86.1\%&89.8\% \\\hline
SGN&96.1\%& 95.5\% \\\hline
\end{tabular}
\caption{Clean accuracy of models trained with standard training~(ST) and adversarial training~(AT) on the HDM05 dataset.  }
\label{tab:at_acc}
\end{center}
\end{table}

\section{Conclusion}
 In this paper, we proposed the first bone length adversarial attack on a skeleton-based action recognition models in an extremely low-dimensional setting. Unlike existing attacks, the proposed attack does not manipulate the motion of skeletons and only perturbs the lengths of the skeleton's bones, the number of which is approximately 30. Nevertheless, for some datasets and settings, it was possible to fool models with small perturbations at a success rate of over 90\%. The skeletons before and after the attack appeared very similar, which makes our attack difficult to notice. We also found that perturbing the bones that were longer and closer to the root joint was more effective. 
 Furthermore, we observed some interesting properties, which are considered to be a characteristic of our low-dimensional setting:   (i) data augmentation improved both clean accuracy and adversarial robustness, and (ii) adversarial training using our attack also improved both of them. To the best of our knowledge, neither of these results have been reported in the standard high-dimensional setting; thus, we consider that our study opens a new direction for adversarial attacks. 
 
\section*{Acknowledgements}
This work was supported by JSPS KAKENHI Grant Number JP19K12039.

\bibliography{aaai.bib}
\end{document}


\maketitle

This supplementary material describes the preprocessing of the HDM05 dataset~\citep{M2007Documentation} and the results of additional experiments. Cross-referencing numbers here are prefixed with S~(e.g., Figure~S1). Numbers without the prefix~(e.g., Figure~1 or Table~1) refer to numbers in the main text. The contents of this supplementary material are summarized as follows.

\section{Preprocessing for the HDM05 dataset}

In the experiments, we used the HDM05 dataset that was preprocessed based on~\citep{Du2015Hierarchical}. 
Our preprocessing can be slightly different from the original one because the details of the preprocessing were not reported in~\citep{Du2015Hierarchical}. 
For reproducibility, we here describe the details of our preprocessing. 

Let the $k$-th sample in the dataset be
\begin{align}
X^{(k)} &= \qty{ \bm{q}_i^{(k)}(t) \mid i=1,\ldots,M, t=0,\ldots,T_k -1},
\end{align}
where $T_k$ and $M$ denote the number of frames of $k$-th sample and joints, respectively, and $\bm{q}_i^{(k)}(t)\in \mathbb{R}^3$ denotes the 3D coordinate of the $i$-th joint of the skeleton at frame $t$. 

First, the Savitzky-Golay smoothing filter~\citep{savitzky1964smoothing} is applied for $t=2,3,\ldots,T_k -3$ as follows.
\begin{align}
\bm{q}_i^{(k)}(t)&=\frac{1}{35} \qty(-3\bm{q}_i^{(k)}(t-2)+12\bm{q}_i^{(k)}(t-1)+17\bm{q}_i^{(k)}(t)+12\bm{q}_i^{(k)}(t+1)-3\bm{q}_i^{(k)}(t+2)). 
\end{align}
Next, we calculate the origin of the coordinates of the joints in frame $t$ for each $k$ as follows.
\begin{align}
\mathcal{O}_k(t) = \frac{1}{3}\qty(\bm{q}_{\mathrm{root}}^{(k)}(t) + \bm{q}_{\mathrm{right}}^{(k)}(t) + \bm{q}_{\mathrm{left}}^{(k)}(t)),
\end{align}
where $\bm{q}_{\mathrm{right}}^{(k)}(t)$ and $\bm{q}_{\mathrm{left}}^{(k)}(t)$ is the right and left joints connected to the root joint, respectively~(i.e., the 17-th and 13-th joints in Fig.~2).
With this origin, each $\bm{q}_{i}^{(k)}(t)$ is translated and redefined by $\bm{q}_{i}^{(k)}(t) - \mathcal{O}_k(t)$. 
Then, we calculate a set of mean vectors $N = \qty{ \bm{\mu}_i \mid i=1,\ldots,M}$ over frames and samples in the training set and validation set. Namely, 
\begin{align}
\bm{\mu}_i = \frac{1}{\sum_{l \in \mathcal{I}_{\mathrm{train}+\mathrm{val}}} T_l} \sum_{k \in \mathcal{I}_{\mathrm{train}+\mathrm{val}}} \sum_{t=0}^{T_k -1} \bm{q}_i^{(k)}(t),
\end{align}
where $\mathcal{I}_{\mathrm{train}+\mathrm{val}}$ denotes the index sets of the union of the training set and validation set. 
The set of standard deviation vectors, $S=\qty{ \bm{\sigma}_i \mid i=1,\ldots,M}$, is calculated in a similar way. Using the means and standard deviations, each sample is normalized as follows.
\begin{align}
\bm{q}_i^{(k)}(t) &= \frac{\bm{q}_i^{(k)}(t) - \bm{\mu}_i}{\bm{\sigma}_i},
\end{align}
where the division by $\bm{\sigma}_{i}$ is performed entry-wise. 
Finally, we sample the frames from $X^{(k)}$ in the four-frame interval to reduce the computational cost and then extended by zero-padding so that the number of frames becomes identical across all the samples. We merged similar classes into a single class as in \citep{cho2014classifying}.

\section{Results with the Adam optimizer}
\begin{figure}[t]
\begin{center}
\includegraphics[width=0.6\columnwidth]{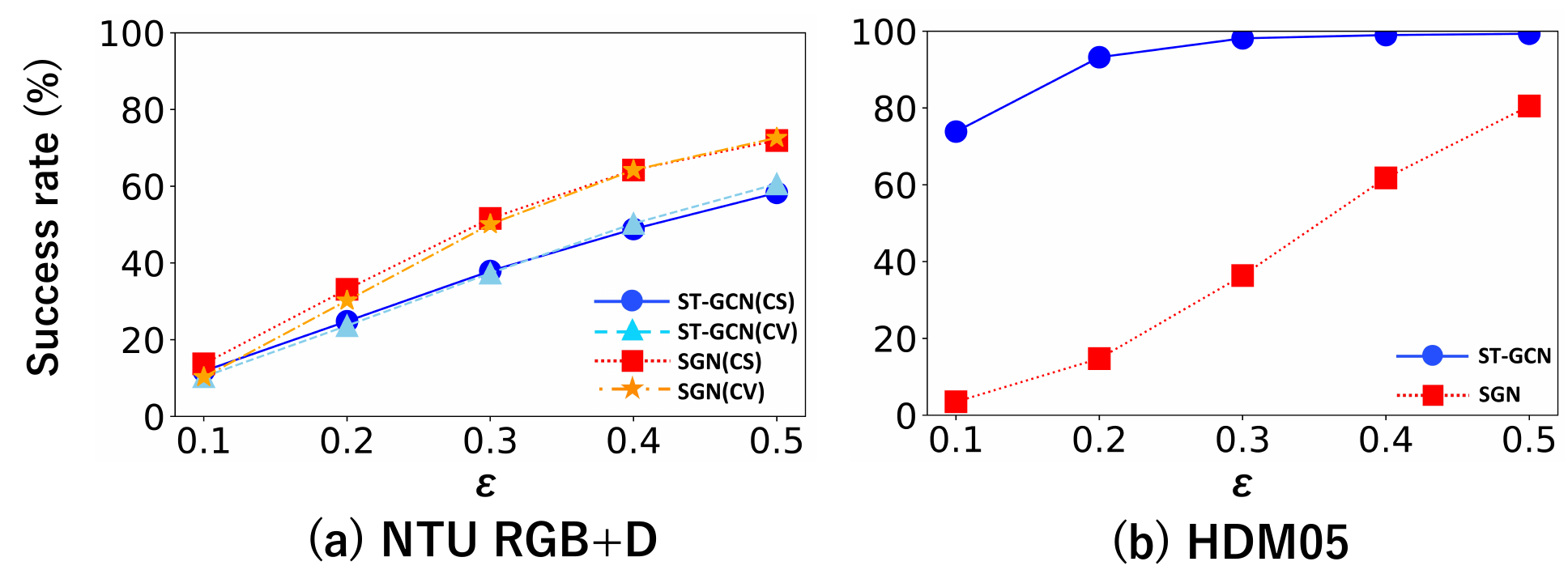}
\end{center}
\caption{Success rates of the adversarial bone length attack using adam on the ST-GCN and SGN models. (a) the NTU RGB+D dataset, (b) the HDM05 dataset.}\label{fig:adam-result}
\end{figure}
Throughout the experiments in the main text, we used PGD~\citep{Madry2018Towards} for the adversarial bone length attack because it is simple and one of the most widely used adversarial attack methods. Some studies on adversarial attacks on skeleton-based action recognition used the Adam optimizer~\citep{Kingma2015Adam} instead of PGD~\citep{Liu2020Adversarial, Wang2021Understanding}. Here, we report the results of our experiments with the Adam optimizer. As shown in Fig.~\ref{fig:adam-result}, the use of the Adam optimizer does not change the main results reported in the main text. With the NTU RGB+D dataset~\citep{Shahroudy2016NTU}, the SGN model~\citep{Zhang2020Semantics} was slightly more vulnerable than the ST-GCN model~\citep{Yan2019Spatial}, while with the HDM05 dataset, the ST-GCN model was extremely vulnerable even for small perturbations. 
